\documentclass{article}

\usepackage{microtype}
\usepackage{booktabs} 
\usepackage{multirow}
\usepackage{hyperref}

\usepackage[accepted]{icml2025}

\usepackage{amsmath}
\usepackage{amssymb}
\usepackage{mathtools}
\usepackage{amsthm}
\usepackage{graphicx}
\usepackage{subcaption}

\usepackage[capitalize,noabbrev]{cleveref}

\theoremstyle{plain}

\theoremstyle{definition}

\theoremstyle{remark}

\usepackage[textsize=tiny]{todonotes}
\usepackage{xcolor} 

\icmltitlerunning{TabularARGN}

\begin{document}

\twocolumn[
\icmltitle{TabularARGN: A Flexible and Efficient Auto-Regressive Framework for Generating High-Fidelity Synthetic Data}



\icmlsetsymbol{equal}{*}

\begin{icmlauthorlist}
\icmlauthor{Paul Tiwald}{comp}
\icmlauthor{Ivona Krchova}{equal,comp}
\icmlauthor{Andrey Sidorenko}{equal,comp}
\icmlauthor{Mariana Vargas Vieyra}{equal,comp}
\icmlauthor{Mario Scriminaci}{comp}
\icmlauthor{Michael Platzer}{comp}
\end{icmlauthorlist}

\icmlaffiliation{comp}{MOSTLY AI, Vienna, Austria}

\icmlcorrespondingauthor{Paul Tiwald}{paul.tiwald@mostly.ai}
\icmlcorrespondingauthor{Michael Platzer}{michael.platzer@mostly.ai}

\icmlkeywords{Machine Learning, Synthetic Data, Data Generation, Tabular Data, Generative AI, Auto-Regressive Generative Neural Network, ICML}

\vskip 0.3in
]



\printAffiliationsAndNotice{\icmlEqualContribution} 

\begin{abstract}
Synthetic data generation for tabular datasets must balance fidelity, efficiency, and versatility to meet the demands of real-world applications. We introduce the Tabular Auto-Regressive Generative Network (TabularARGN), a flexible framework designed to handle mixed-type, multivariate, and sequential datasets. By training on randomized subsets of conditional probabilities, TabularARGN supports advanced features such as fairness-aware generation, imputation, and conditional generation on any subset of columns. The framework achieves state-of-the-art synthetic data quality while significantly reducing training and inference times, making it ideal for large-scale datasets with diverse structures. Evaluated across established benchmarks, including realistic datasets with complex relationships, TabularARGN demonstrates its capability to synthesize high-quality data efficiently. By unifying flexibility and performance, this framework paves the way for practical synthetic data generation across industries.
\end{abstract}


\section{Introduction}
\label{sec:introduction}

Equitable and broad access to data is essential for advancing research, driving innovation, and addressing pressing societal challenges \cite{hradec2022multipurpose}. Despite the abundance of valuable datasets within public and private organizations, much of this information remains inaccessible due to privacy concerns and the limitations of traditional anonymization techniques, which often fail to prevent re-identification in high-dimensional data \cite{gadotti2024anonymization}. Securely unlocking such data can yield substantial benefits, enabling progress both within organizations and across society. Synthetic data has emerged as a promising solution to this challenge, offering a means to securely share and analyze sensitive datasets by preserving their analytical utility while mitigating privacy risks \cite{drechsler2011synthetic, jordon2022synthetic, united2023synthetic, hu2023sokprivacypreservingdatasynthesis}.

By leveraging deep neural network models to learn the patterns, distributions, and relationships within original data, synthetic samples can be generated that are structurally consistent, statistically representative, and truly novel, thereby effectively minimizing disclosure risks. Importantly, the benefits of synthetic data extend beyond privacy protection, as the fitted generative models enable the creation of arbitrary data volumes, rebalancing of underrepresented groups, imputation of missing data points, scenario-specific conditional sampling, and adherence to statistical fairness \cite{NEURIPS2021_ba9fab00}. These capabilities significantly enhance the analytical value and practical usability of datasets, supporting a wide range of downstream applications and driving meaningful progress across diverse domains \cite{vanbreugel2023privacynavigatingopportunitieschallenges}.

The core of synthetic data generation is \textit{robust} density estimation: accurately approximating the joint distribution of multivariate data while minimizing the influence of individual training samples. Synthetic data shall closely resemble original holdout samples drawn from the same distribution, without being any closer to the training data than holdouts are \cite{pla21-qa}. By sampling from the estimated distribution, synthetic data preserves the statistical relationships of the original dataset, achieving high fidelity and strong privacy safeguards.

In this paper, we revisit auto-regressive models for density estimation, which approximate the full joint density $p(\mathbf{x})$ as a product of conditional probabilities. This decomposition leverages the chain rule of probability, where $p(\mathbf{x}) = \prod_{i=1}^D p(x_i \mid x_1, \ldots, x_{i-1})$, breaking a complex density estimation problem into a series of simpler conditional distributions. Auto-regressive models have proven remarkably successful in density estimation tasks in natural language processing \cite{brown2020languagemodelsfewshotlearners, openai2024gpt4technicalreport} and image generation \cite{tian2024visualautoregressivemodelingscalable, sun2024autoregressivemodelbeatsdiffusion}, underpinning recent advances in synthetic image creation and, more prominently, the transformative success of large language models (LLMs).

While there is a growing body of literature leveraging LLM-type models \cite{hegselmann2023tabllm, smolyak2024large, miletic2024}, both token-based custom-built transformer architectures and fine-tuned foundational language models for tabular synthetic data generation, these approaches often treat rows of tabular data as sequences of text. In contrast, the literature on specialized auto-regressive models designed specifically for tabular data, in its various forms and structures, remains scarce. These specialized models do not rely on treating rows as text but instead exploit the constrained value ranges of structured tabular datasets. Motivated by this gap, we introduce the Tabular Auto-Regressive Generative Network (TabularARGN) framework, which applies auto-regressiveness across the discretized column, time, and table dimensions to generate high-fidelity synthetic data for flat and sequential tables.

We define flat tables as datasets where each row represents an independent and identically distributed (i.i.d.) sample from an underlying distribution. In the context of privacy protection, each data subject, whose privacy is to be safeguarded, is represented by a single row. Sequential tables, by contrast, contain multiple records per data subject, grouped by a unique key provided within the tabular data. Furthermore, our presented framework supports the inclusion of additional context for each training sample through a separate table associated via a primary–foreign key, one-to-many relationship. For example, a flat table could include time-independent covariates, such as a bank customer’s date of birth or membership status, while the sequential table captures the customer’s transaction history.

TabularARGN enables the generation of multivariate, mixed-type synthetic data for both flat and sequential tables. The sequential samples are generated conditionally on their respective subjects in the flat table, ensuring consistency between the two. TabularARGN is designed to operate on multi-sequence data, where each data subject has an associated sequence in the sequential table. As these sequences often have varying lengths, TabularARGN is also equipped to estimate and synthesize the sequence-length distribution observed in the original data, ensuring a realistic representation of sequential structures.
\section{Related work and contributions}
\label{sec:related_work}

To the best of our knowledge, \citet{uri16} were the first to develop an auto-regressive neural network (NADE) for tackling the problem of unsupervised distribution and density estimation. This foundational idea and architecture serve as the backbone of the TabularARGN flat and sequential models.

\subsection{Flat Models}
In their study, \citet{uri16} calculate the log-likelihood performance of NADE on binary tabular and binary image datasets, as well as purely real-valued datasets. However, they do not extend their work to generate synthetic data samples. Furthermore, the NADE architecture is not adapted in their work to handle multi-categorical datasets with variables of cardinality $\geq 2$, nor mixed-type datasets that combine categorical and numeric variables, both of which are critical for real-world tabular data synthesis tasks. 
NADE shares similarities with Bayesian Networks (BNs), as both methods factorize the joint probability into conditional distributions. 
However, BNs represent dependencies between variables using a directed acyclic graph. 
Once the conditional probability distributions are learned, data can be sampled from the model, making BNs a natural choice for generating tabular data \cite{ank15, qia23}.



Variational Autoencoders (VAEs) \cite{kingma13-vae} and Generative Adversarial Networks (GANs) \cite{goo14-gan} represent two prominent classes of generative models that are frequently used for synthetic data generation. VAEs explicitly model joint distributions by learning a latent representation of the data, while GANs implicitly approximate these distributions through an adversarial training process between a generator and a discriminator. Both architectures have been adapted to handle mixed-type tabular synthetic data [VAE-based: \cite{akr22-VAE, liu23-goggle}; GAN-based: \cite{par18, xu19, zha21, qia23, li23-ehrMGAN, zha24}].


With the rise in popularity of Large Language Models (LLMs), token-based transformers have also been applied to generate tabular synthetic data based on a specific target dataset. Some methods finetune pre-trained LLMs directly, while others train LLM-like architectures from scratch \cite{bor23, sol23-Realtabformer, kar24synehrgy}. 

In addition to token-based transformers, auto-regressive transformer models specifically designed for tabular data have also been proposed \cite{led21, cas23, gul23-TabMT}. Unlike LLMs or token-based approaches, these models explicitly leverage the inherent structure of tabular data, such as limited value ranges, column-specific distributions, and inter-feature relationships, to enhance both the efficiency and quality of synthetic data generation. More recently, researchers have begun exploring hybrid approaches that combine transformers with diffusion to model discrete features \cite{zha24-TabDar}.

Diffusion models, originally developed for image and audio synthesis, have also been adapted for generating tabular synthetic data \cite{lee23-Codi, kim23-StaSy, kot23-TabDDPM, zha24-TabSyn, vil24}. These models leverage iterative denoising processes to approximate complex data distributions and have shown promise in capturing intricate patterns in tabular datasets. It has also been shown that diffusion models can effectively memorize the training data with a larger number of training epochs \cite{fang2024understandingmitigatingmemorizationdiffusion}.

There is an ongoing discussion in the machine learning community about why tree-based methods often outperform neural networks when applied to tasks involving tabular data such as prediction \cite{gri22-NNvsTree}. This observation has inspired the development of tree-based generative models for tabular data \cite{jol24-ForrestDiffusion}. Some models typically train per-feature models and employ them in an auto-regressive manner during data generation \cite{mcc24-UnmaskingTrees}.

\subsection{Sequential Models}
The generation of sequential data, very much tied to time-series generation and continuation \cite{yoo19-TimeGAN, lin20-DoppelGANger, des21-TimeVAE, zhi24-sdformer, yua24-diffusionts, suh24-timeautodiff}, has been extensively studied in the literature, resulting in numerous models and methods for time-series synthesis. However, many of these implementations are not directly applicable for benchmarking against complex real-world datasets. Common limitations include the inability to process discrete features, reliance on single-sequence settings with sequences chunked during training, restrictions to constant sequence lengths across samples, reliance on constant or equidistant time steps, and a lack of support for missing values. As a result, our focus is on frameworks that, like TabularARGN, can synthesize mixed-type, multivariate, multi-sequence data with non-trivial sequence-length distributions, and support conditioning on a flat table to reflect realistic use cases \cite{pat16-sdv, sol23-Realtabformer, gue23-rctgan, pan24-clavaddpm}.

\subsection{Our Contributions}

In this work, we introduce the Tabular Auto-Regressive Generative Network (TabularARGN) framework for synthetic data generation. Departing from the trend of increasingly large and complex models, TabularARGN adopts a simple yet effective architecture built on well-established principles. The framework uniquely delivers the following key contributions:

\begin{itemize}
    \item \textit{High Fidelity}: TabularARGN achieves synthetic data quality on par with state-of-the-art (SOTA) models.
    \item \textit{Privacy by Design}: TabularARGN only considers privacy-preserving value ranges for sampling, has built-in regularization layers, applies early stopping for training, plus can be trained via DP-SGD for obtaining differential privacy guarantees \cite{dwo14-dp, aba16-dpsgd}.
    \item \textit{Compute Efficiency}: The training and sampling of TabularARGN models are sometimes orders of magnitude faster than other generative SOTA models for tabular data, making it possible to apply these to significantly larger real-world datasets.
    \item \textit{Simplicity}: TabularARGN leverages existing building blocks, and thus can be easily implemented within standard deep learning frameworks.
    \item \textit{Sampling Flexibility}: 
        TabularARGN supports conditional generation on arbitrary subsets of mixed-type attributes enabling data generation tailored to specific constraints. It also facilitates missing value imputation and the incorporation of statistical fairness across sensitive attributes \cite{krc23}. Additionally, sampling probabilities can be adjusted via a temperature parameter to balance rule adherence with data diversity.
    \item \textit{Data Versatility}: TabularARGN accommodates the heterogeneity of real-world tabular datasets, including multi-variate, mixed-type data (categorical, numerical, date-time, geospatial), multi-sequence datasets with varying sequence lengths and varying time intervals, and missing values.
    \item \textit{Robustness in Training}: TabularARGN delivers high-quality synthetic data with default settings and remains consistent across several training runs.
    \item \textit{Open-Source Framework}: We provide a well-tested and well-maintained implementation of TabularARGN under a fully permissive Apache v2 Open Source license at \url{https://github.com/mostly-ai/mostlyai-engine/}.
\end{itemize}

By addressing these aspects in combination, TabularARGN aims to bridge the gap between synthetic data research and real-world applications, enabling robust performance in the complex, uncontrolled environments typical of industry-scale data systems.
\section{The TabularARGN framework}
\label{sec:architecture}

TabularARGN is a shallow any-order auto-regressive network architecture built upon discretized attributes, and trained for minimizing the categorical cross-entropy loss.

Key aspects are 1) the simultaneous multi-target training, 2) the consistent robust loss across mixed-type columns, 3) the constrained sampling domain given the original value ranges, and 4) the any-order auto-regressive setup, which allows for full flexibility during sampling.

\subsection{Data Encoding}

Prior to training, the original data is analyzed to derive a privacy-safe domain of value ranges. For categorical columns, any rare categorical values are mapped into a single category, to protect against basic membership inference attacks based on sampled value ranges. For numerical columns, values are either binned and mapped onto percentiles or are split into individual digits. For datetime columns, the values are split into their date and time components, i.e. year, month, day, etc. Both for numeric and datetime data, the value ranges are clipped, to be insensitive towards the inclusion of individual outliers (e.g. a single person of age 123 years). For geospatial data, the combination of latitude and longitude is mapped into a sequence of quadtiles, with its resolution automatically adapting to the data density within a given region.

Ultimately, every column, independent of its original data type, will be mapped onto one or more categorical sub-columns as part of the encoding phase. This mapping increases the number of features. In the remainder of the paper, $D$ will denote the number of sub-columns. During sampling, that mapping is then reversed to decode any sampled data back into the original data domain. See Appendix \ref{A:encoding} for details on the encoding strategies.

\subsection{The Flat Table Model}
The Tabular ARGN model is designed to generate tabular synthetic data by learning the full set of conditional probabilities across features in a dataset. At its core, the model maps the generation process to an auto-regressive framework, estimating the joint probability distribution of the data as a product of simpler conditional probabilities. For each feature $x_i$, the model learns to output the estimated discrete conditional probabilities $\hat{p}(x_i \mid x<i)$, where $x<i$ represents the set of preceding features.

During training, TabularARGN incorporates a flexible "any-order" approach, wherein the order of features is dynamically shuffled for each training batch. This allows the model to estimate not only "fixed-order" conditional probabilities $\hat{p}(x_i \mid x<i)$ present in the input data set, but also probabilities conditioned on any subset of features:
\begin{equation}
    \forall S \subseteq \{1,\dots,D\} \setminus \{i\}, \quad \hat{p}(x_i \mid \{x_j : j \in S\}).
\end{equation}
This flexibility ensures that the model can adapt to arbitrary auto-regressive conditioning scenarios during generation, a \textit{conditio sine qua non} for applications such as imputation, fairness adjustments, and conditional generation.

A similar order-agnostic approach was already suggested by \citet{uri16} within the NADE framework, adapted to the natural-language processing domain \cite{yan19-xlnet,shih2022}, and is also discussed in more recent implementations of flat table synthetic data generation \cite{led21, gul23-TabMT, mcc24-UnmaskingTrees}.

\begin{figure*}
     \centering
     \begin{subfigure}[b]{0.45\textwidth}
         \centering
         \includegraphics[width=\textwidth]{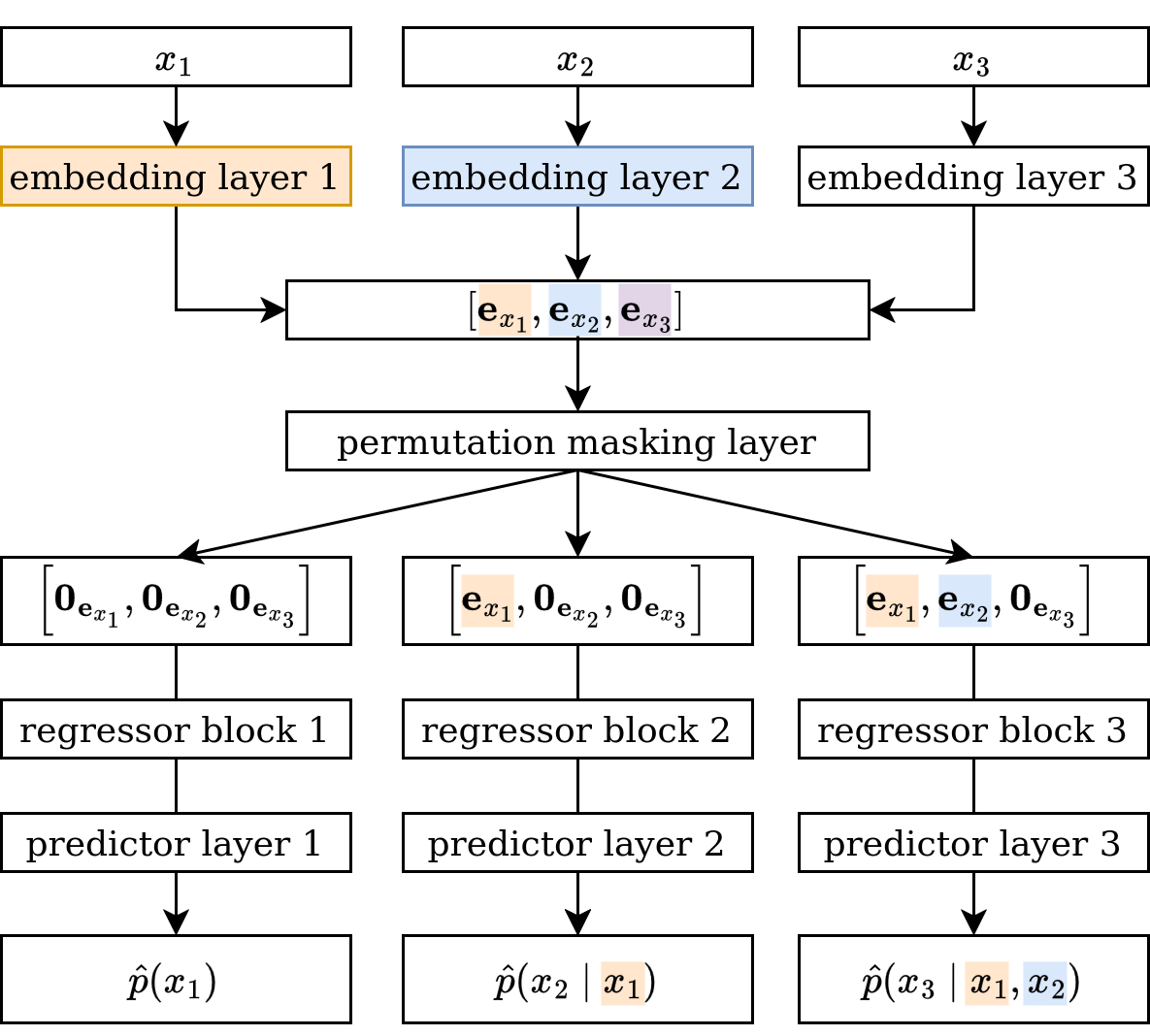}
         \caption{}
         \label{img:flat-model-train}
     \end{subfigure}
     \hfill
     \begin{subfigure}[b]{0.45\textwidth}
         \centering
         \includegraphics[width=\textwidth]{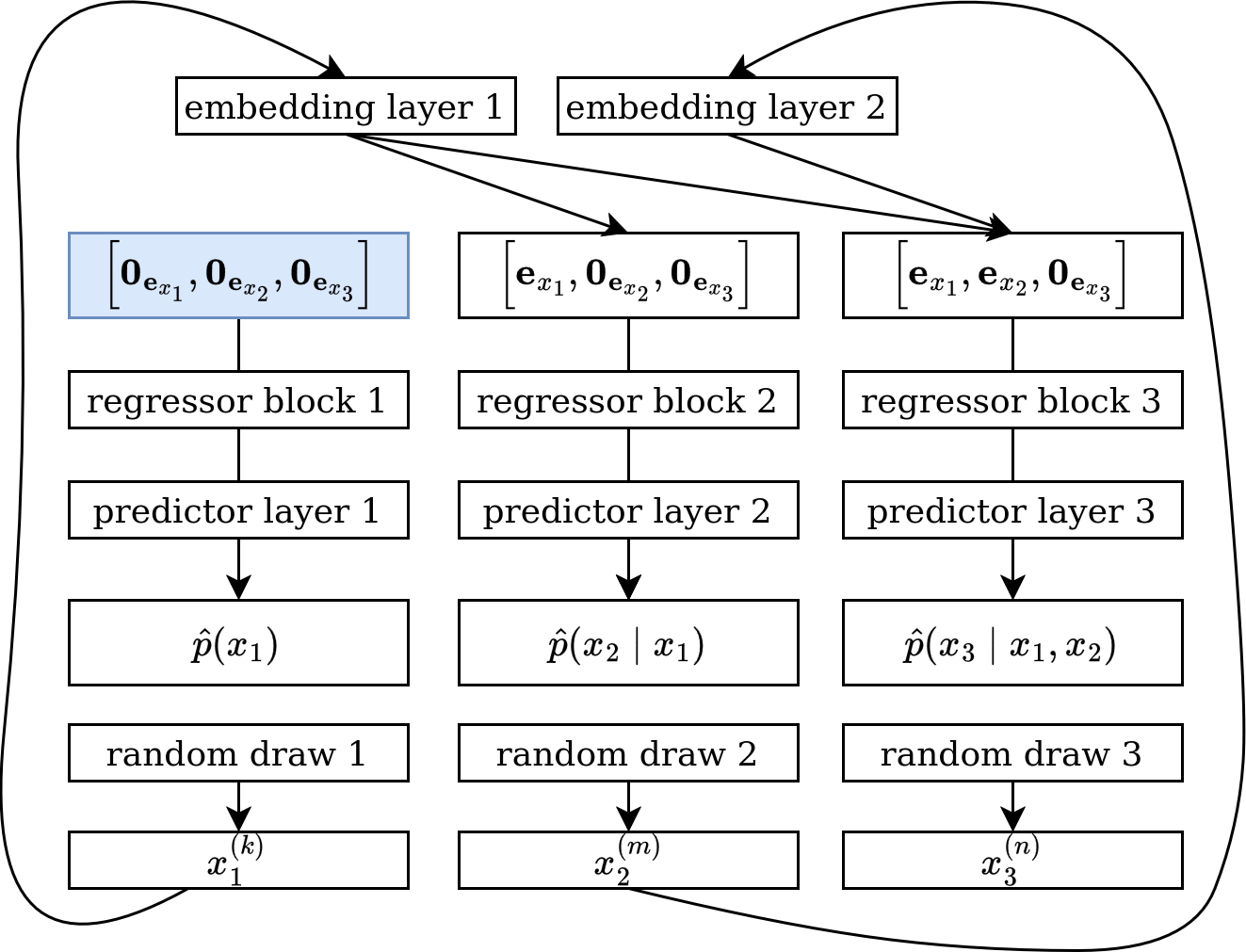}
         \caption{}
         \label{img:flat-model-gen}
     \end{subfigure}
     \caption{(a) Model components and information flow in the \textbf{training phase} of a three-column TabularARGN flat model with the current column order [1,2,3]. Input features $x_{i}$ are embedded and sent through the permutation masking layer to condition predictions on preceding columns. The permutation masking layer randomly shuffles the column order for each training batch. (b) Model components and information flow in the \textbf{generation phase}. The input to the model and starting point of the generation is a vector of zeros (blue) triggering the successive generation of synthetic features. Due to the permutation of column orders during training, any column order can be realized in the generation phase.}
     \vskip -0.2in
\end{figure*}

\subsection{Model Components}
For every sub-column in the dataset, the model incorporates an embedding layer, a regressor block, and a predictor layer, which are interconnected via a permutation masking layer that directs the flow of information (see fig.\ \ref{img:flat-model-train}). 

The trainable embedding layers transform the categories of a sub-column $x_i$ into embedding vectors, denoted as $\mathbf{e}_{x_i}$. The dimensionality of these embedding vectors is determined via a heuristic, that depends on the cardinality of a sub-column, allowing the model to adapt to the complexity of each feature (see appendix \ref{A:heuristics} for the heuristic determining the embedding sizes).

The embedding vectors $\mathbf{e}_{x_i}$ for all sub-columns are concatenated into a single vector $\left[\mathbf{e}_{x_1}, ..., \mathbf{e}_{x_N}\right]$, which is then passed to the permutation masking layer. The permutation layer introduces a "causal" feeding mechanism based on a random column order - varying from batch to batch - ensuring that each column’s regressor receives only the inputs it is allowed to access based on the conditional dependencies. The masking layer modifies the concatenated vector by "blanking out" (i.e., setting to zero) the contributions not permitted for specific regressor layers.

In a three-column example (see fig.\ \ref{img:flat-model-train}), the regressor for column $x_1$ receives a vector of zeros, $\mathbf{0}_{\left[\mathbf{e}_{x_1}, \mathbf{e}_{x_2}, \mathbf{e}_{x_3}\right]}$, with dimensions matching $\left[\mathbf{e}_{x_1}, \mathbf{e}_{x_2}, \mathbf{e}_{x3}\right]$. The regressor for column $x_2$ receives $\left[\mathbf{e}_{x_1}, \mathbf{0}_{\mathbf{e}_{x_2}}, \mathbf{0}_{\mathbf{e}_{x_3}}\right]$, where $\mathbf{0}_{\mathbf{e}_{x_2}}$ and $\mathbf{0}_{\mathbf{e}_{x_3}}$ represent zero vectors of sizes equal to $\mathbf{e}_{x_2}$ and $\mathbf{e}_{x_3}$, respectively. The regressor for column $x_3$ receives $\left[\mathbf{e}_{x_1}, \mathbf{e}_{x_2}, \mathbf{0}_{\mathbf{e}_{x_3}}\right]$, allowing it to condition on the embeddings of both preceding columns.

The regressor block consists of one or more feed-forward layers (see appendix \ref{A:heuristics} for details), each using a ReLU activation function. To enhance generalization and prevent overfitting, dropout is applied during training to each layer within the regressor block. The predictor layer is a single feed-forward layer with a softmax activation applied during generation, outputting normalized, discrete probability vectors that represent the estimated probabilities for each category $x_i^{(k)}$. The size of the predictor layer corresponds to the number of categories within the specific sub-column.

\subsection{Model Training}

The training target of TabularARGN is the minimization of the categorical cross-entropies, computed and summed up across each sub-column. With the "any-order" permutations, this procedure effectively minimizes the negative log-likelihood
\begin{equation}
    \max_{\theta} \;
    \mathbb{E}_{\sigma \in \text{Uniform}(S_D)}
    \Biggl[
    \sum_{i=1}^{D}
    \log p_{\theta}\bigl(x_{\sigma(i)} \mid x_{\sigma(<i)}\bigr)
    \Biggr],
\end{equation}
where $\sigma$ is a random permutation, uniformly drawn from the symmetric group $S_D$, with $D$ being the number of features (sub-columns). $x_{\sigma(<i)}$ denotes all features that precede feature $i$ in the permutation $\sigma$.

During training, teacher forcing is employed, where ground-truth values from preceding columns are provided as inputs to condition the model. This approach effectively treats the training process as a multi-task problem, wherein each sub-column represents a distinct predictive task.

The training procedure incorporates a robust stopping mechanism to ensure efficient convergence. A validation set is split from the input dataset, and the validation loss is scored after every epoch. A patience mechanism is employed, which begins to reduce the learning rate once the validation loss ceases to improve. Training is stopped when further learning rate reductions fail to yield additional improvements in validation loss. The model weights corresponding to the lowest validation loss are retained, and this state is referred to as the converged model or the convergence of model training. Details of the patience mechanism are provided in appendix \ref{A:model-training}.


From an architectural perspective, it is possible to train separate, independent models for each sub-column. This per-column approach is commonly employed in tree-based (i.e., non-deep-learning) methods for generating tabular synthetic data, where models for individual columns are trained sequentially and conditionally based on preceding columns \cite{mcc24-UnmaskingTrees}. However, our empirical observations show that when training such individual, per-column models to convergence, the summed validation losses of these independent models consistently exceed the converged validation loss of the multi-task model. This behavior suggests that using shared embeddings across columns confers an advantage by capturing and transferring relationships across different predictive tasks, improving overall model generalization.

\subsection{The Sequential Table Model}

The TabularARGN model for sequential tables is auto-regressive along both the column dimension (as the flat model) and the time/sequence dimension. This means that, in addition to estimating the discrete probability distribution of a column conditioned on the values of previous columns within the same time step, the column-specific regressors of sequential table models also ingest the encoded history of all previous time steps (see blue part in fig.\ \ref{img:sequential-model-train}). For example, when generating values for the $j$-th sub-column for time step $t=T$ of a sequential table, the model (in the fixed column-order setting) learns to estimate the conditional probability distribution $\hat{p}_{x_j, t=T}(x_j \mid x_{i<j, t=T}, \mathcal{H}_{0:T-1})$, where $\mathcal{H}_{0:T-1}$ denotes the encoded history of all previous time steps.

The historical encoding $\mathcal{H}_{0:T-1}$ is provided by the last hidden state of a history encoder consisting of an LSTM (Long Short-Term Memory) layer \cite{hoc97-lstm} - with dropout during training - which captures temporal dependencies and patterns in the data (see appendix \ref{A:heuristics} for the size of the latent history representation). The input vector to the history encoder is $\left[\mathbf{e}_{x_1; t=0:T-1}, \mathbf{e}_{x_2;t=0:T-1}, \mathbf{e}_{x_3; t=0:T-1}\right]$, the concatenated embeddings after the application of a temporal causal shift. Beginning and end time steps, $t=0$ and $t=T-1$, indicate that the embedding vectors are padded with zeros at $t=0$ and the last input time step $t=T$ is dropped. $\mathcal{H}_{0:T-1}$ is concatenated with the column embeddings of previous columns $\left[\mathbf{e}_{x_{i<j}; t=T}, \mathcal{H}_{0:T-1} \right]$, and fed as input to the regressor of column $j$.

The training of sequential table TabularARGN models also relies on the categorical cross-entropy loss function. However, unlike flat table models, where the total training loss is summed across sub-columns, sequential table models extend this summation across both columns and time steps. This approach ensures that the training process accounts for temporal dependencies while still treating each sub-column at each time step as a distinct predictive task within the multi-task framework.

The sequential table TabularARGN model can handle sequences of arbitrary lengths during both training and generation. Moreover, it does not require sequence items or time steps to be equidistant, nor does it require any special columns containing time step information in a specific format. In other words, the model can handle not only sequences of arbitrary lengths (including "zero" sequence length) but also "irregular" sequences, making it suitable for a wide range of real-world applications. Further details are provided in appendix \ref{A:sequential-table-model}. This flexibility allows the sequential table model to train on and generate not only time-series data but also (ordered) sets.

The any-order logic, as described for the flat table model, is equally applicable to the column dimension of the sequential table model. This flexibility allows for adaptive conditioning and context-driven generation, enabling TabularARGN to handle a wide range of sequential data scenarios efficiently and with high fidelity.

\subsection{Conditional Sequence Generation - Two-Table setups}

In addition to conditioning on column values and time-series history, TabularARGN’s sequential table model can utilize a flat table as context during training and generation, enabling the synthesis of two-table setups, such as a flat table containing time-independent information (e.g.\ bank customer data) and a sequential table containing time-dependent data (the bank customers' transaction histories).

\begin{figure*}
     \centering
     \includegraphics[width=0.8\textwidth]{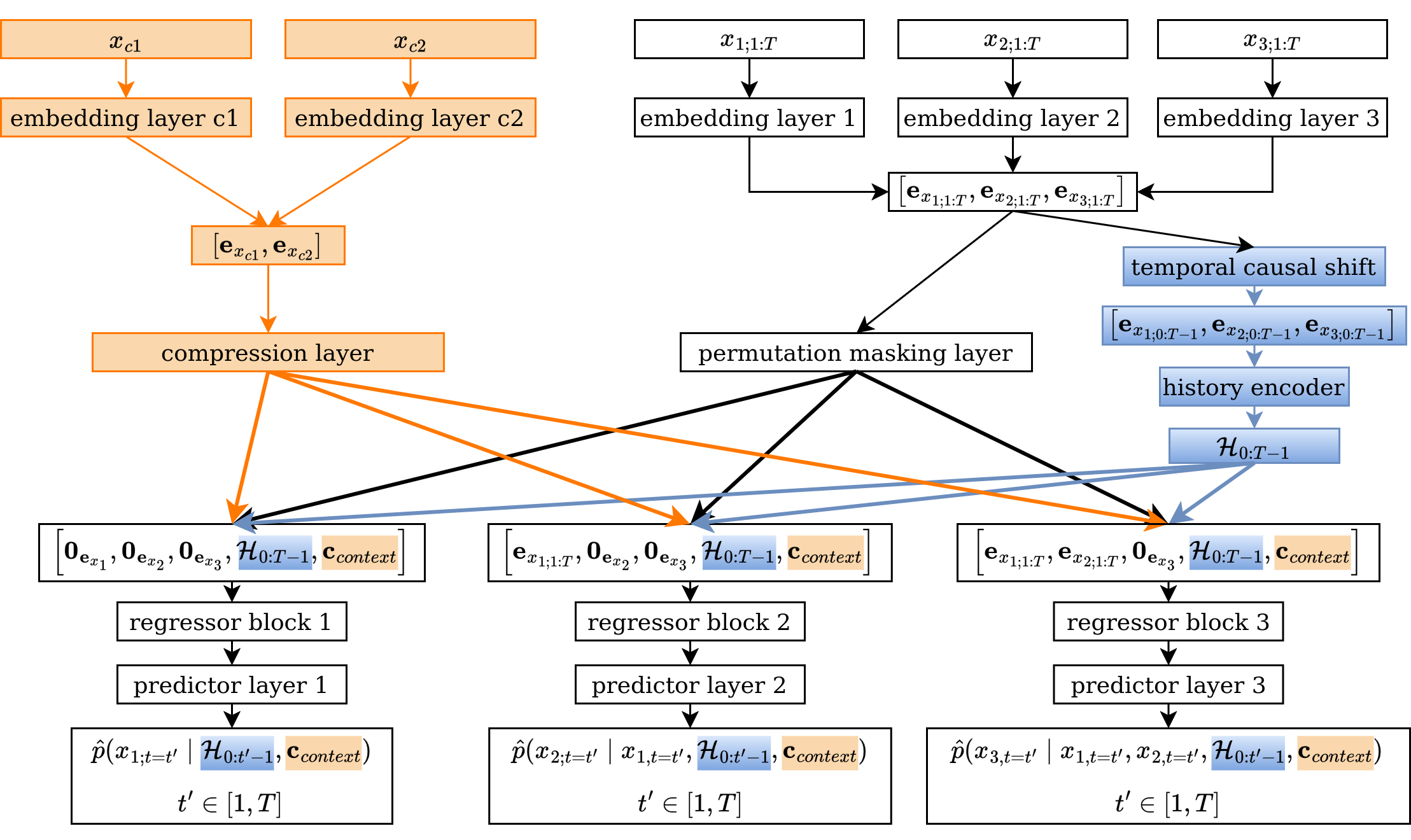}
     \caption{Model components and information flow in the training phase of a three-column TabularARGN sequential model with a two-column flat context. The current column order is [1,2,3]. Input features $x_{i;1:T}$ are embedded and sent through both the permutation masking layer and the history encoder block (blue) to condition predictions on preceding columns and time steps, respectively. For conditional sequence generation, flat context features $x_{ci}$ are ingested by the context processor (orange). They are encoded, compressed, and provided as context for the prediction of all columns of the sequential training data.}
     \label{img:sequential-model-train}
     \vskip -0.2in
\end{figure*}

This functionality is facilitated by the context processor module (see orange part in fig.\ \ref{img:sequential-model-train}, which ingests the flat table and produces an additional context embedding $\mathbf{c}_{\text{context}}$. This embedding is concatenated with the column and history embeddings of the sequential table before being passed to the regressor layers, allowing the model to capture correlations between the flat and sequential tables. Specifically, the input fed to the regressor for the sequential table sub-column $j$ at time step $t = T$ is $\left[\mathbf{e}_{x_{i<j}; t=T}, \mathcal{H}_{0:T-1}, \mathbf{c}_{\text{context}}\right]$. The context processor generates $\mathbf{c}_{\text{context}}$ by concatenating individual column embeddings and passing the resulting vector through a feed-forward layer with a ReLU activation function and dropout during training. We provide details on the layer sizes in appendix \ref{A:heuristics}.

In a two-table setup, the training of flat and sequential tables can be performed in parallel, independent of each other. However, during generation, the flat table must first be sampled to provide the necessary context for the sequential table model. In this sense, the auto-regressive approach is extended to the table dimension, complementing its application across the column and time dimensions.


\section{Empirical Results}
\label{sec:results}

In this section and Appendix~\ref{A:further-results}, we compare the synthetic data quality as well as the compute efficiency of TabularARGN to available state-of-the-art (SOTA) models, across a variety of small and medium-scaled datasets. For maintaining focus, we present results for one representative flat dataset and one sequential dataset here, while additional results on additional datasets - demonstrating similarly strong performance of TabularARGN - are provided in Appendix~\ref{A:further-results}. Analogous to the previous sections, we split this section into a study on flat and sequential tables. For each, we utilize selected methods and data sets from existing benchmarks in the literature \cite{zha24-TabSyn, pan24-clavaddpm}.

We evaluate the synthetic data quality of all methods following the evaluation framework proposed by \citet{pla21-qa} and implemented in \citet{mostlyai-qa}. This approach groups evaluation metrics into low-dimensional statistical distances and nearest-neighbor-based privacy metrics. The low-order statistics are measured as discretized univariate and bivariate empirical marginal distributions for flat tables. For sequential scenarios, the auto-correlation of columns within a sequence is also analyzed as a measure of the coherence in the generated sequences. Privacy metrics, expressed as DCR share, measure the proportion of synthetic samples that are closer to a training sample than to a holdout sample. Given that the training and holdout sets are of equal size, a DCR share of 50\% indicates that synthetic samples are, on average, equally "distant" from both sets. This provides empirical evidence for the privacy safety of the synthetic data by assessing the risk of sensitive information being leaked into the generated samples. Detailed descriptions of all metrics can be found in Appendix \ref{A:metrics}. 

In this section, we present results in terms of runtime and synthetic data quality - the overall accuracy computed as the average across univariate, bivariate, and coherence accuracy. We report the DCR shares only in Appendix~\ref{A:further-results} as none of the methods displays a significant increase beyond 50\%. 

\subsection{Flat Tables}
\label{sec:results-flat}

\begin{figure*}
     \centering
     \includegraphics[width=0.95\textwidth]{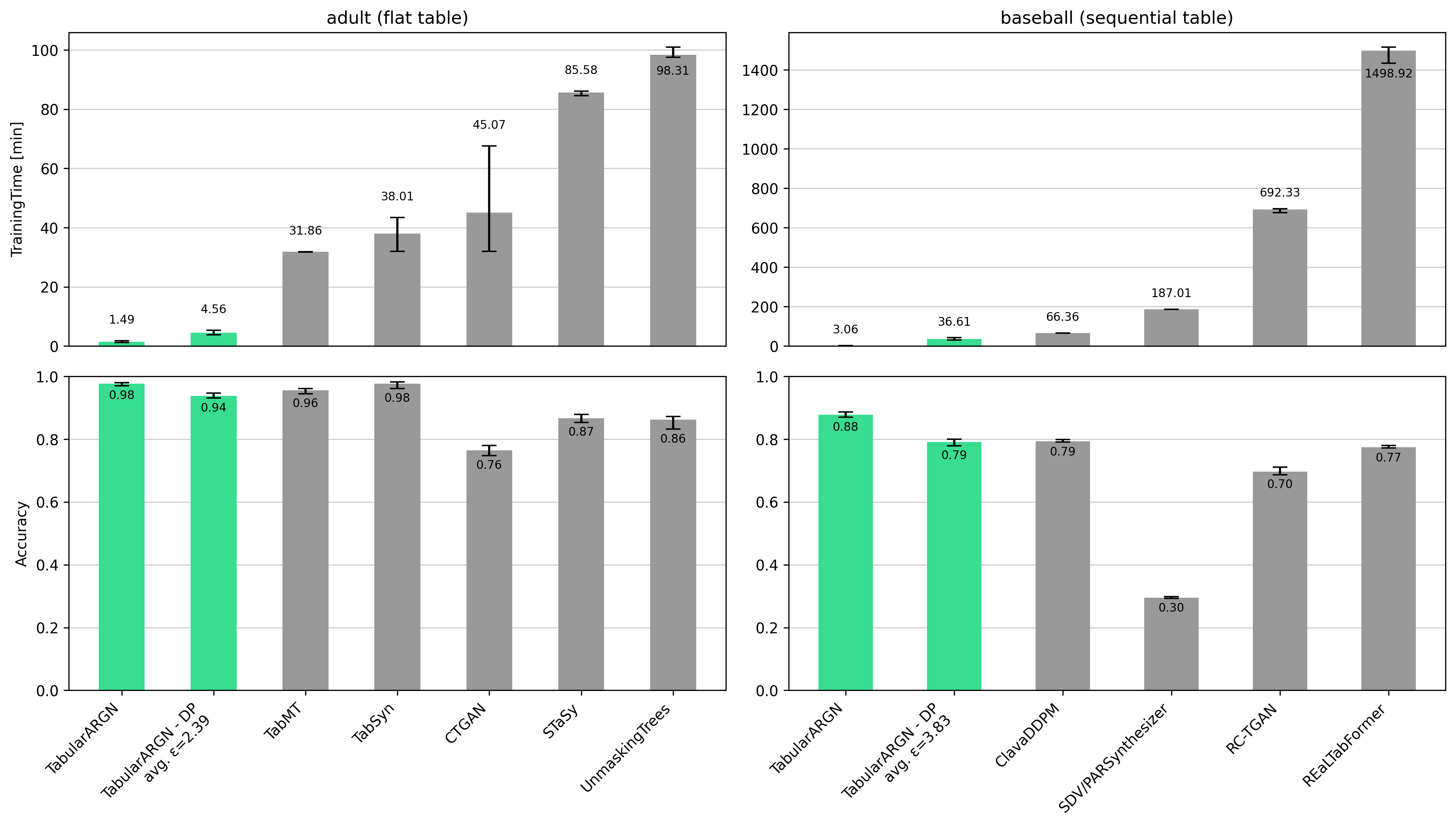}
     \caption{Training time (top row) and Accuracy (bottom row) for the flat \emph{Adult} (left column) and sequential \emph{Baseball} (right column) data set. Reported values are averages over five full training and generation runs, error bars indicate the minimum and maximum values. As UnmaskingTrees is a non-Deep Learning approach, it is run on CPU. Training of TabularARGN on the \emph{Adult} with a CPU takes approximately 2.5 minutes.}
     \label{fig:results}
     \vskip -0.2in
\end{figure*}

As baseline models, we include CTGAN \cite{xu19}, STaSy \cite{kim23-StaSy}, TabSyn \cite{zha24-TabSyn}, TabMT \cite{gul23-TabMT}, and Unmasking Trees \cite{mcc24-UnmaskingTrees}, providing a strong mix of state-of-the-art models from different generative approaches, including a GAN-based method as well as diffusion and auto-regressive models.  We refer the reader to Appendix~\ref{A:experimental-setup} for a more detailed description of these methods, as well as the statistics of all datasets utilized in this work.
On the \emph{Adult} data set \cite{dua19-adult} - 24k rows with eight categorical and six numerical variables - TabularARGN performs on par with state-of-the-art methods such as TabSyn in terms of accuracy while outperforming all benchmarks in terms of runtime by a large margin (see Figure~\ref{fig:results} left column). Notably, when employing Differential Privacy ($\epsilon \sim 2.5$, $\delta=10^{-5}$), performance remains competitive, with only a modest decline in accuracy, while the runtime increases moderately but remains well below that of the baselines. The reported epsilon values correspond to the privacy levels reached at model convergence. Further results on other methods and flat table datasets can be found in Appendix~\ref{A:further-results}.




\subsection{Sequential Tables}

Establishing a comprehensive benchmark for sequential tables presents challenges, as many methods proposed in the literature for sequence generation are constrained by the properties of the data they can handle. As already mentioned earlier, common limitations include the inability to process discrete features, reliance on single-sequence settings with sequences chunked during training, restrictions to constant sequence lengths across samples, and reliance on constant or equidistant time steps. These limitations make it difficult for such methods to handle real-world datasets, which often exhibit diverse structures, varying sequence lengths, and a mix of feature types. The baseline methods selected for our study address all these limitations - at least after some basic pre-processing - and are capable of producing synthetic data of competitive quality. We evaluate REalTabFormer \cite{sol23-Realtabformer}, RC-TGAN \cite{gue23-rctgan}, and ClavaDDPM \cite{pan24-clavaddpm} which represent a diverse cross-section of generative approaches.

We compare TabularARGN with the baseline models on a two-table subset (\textit{players} and \textit{fielding}) of the \emph{Baseball} dataset \citep{lahman2023baseball}. The \textit{players} table (flat context) has about 20 thousand rows, and the sequential \textit{fielding} table has 140 thousand, resulting in an average sequence length of seven. More detailed information on methods and data sets can be found in appendix~\ref{A:experimental-setup}.

TabularARGN achieves the highest accuracy, surpassing all other methods by nearly 10 percentage points (see Figure~\ref{fig:results} right column). Even with differential privacy enabled ($\epsilon\sim4$, $\delta=10^{-5}$), TabularARGN remains on par with the top-performing baseline methods. Even though adding privacy guarantees increases the training time by a factor of 10, our method remains one to two orders of magnitude below those of the baseline models, reinforcing its scalability and practical usability for real-world applications. Further results on other methods and sequential table datasets can be found in Appendix~\ref{A:further-results}.

\section{Conclusion}
In this paper, we introduced the Tabular Auto-Regressive Generative Network (Tabular ARGN), a novel framework for generating high-quality synthetic data for both flat and sequential tables, and share a well-tested and actively maintained open-source implementation. Using a unified autoregressive modeling approach, TabularARGN learns all conditional probabilities across features and time steps, enabling unprecedented flexibility in downstream tasks such as imputation, fairness adjustments, and group-specific conditional generation.

Our results demonstrate that TabularARGN achieves synthetic data quality on par with state-of-the-art methods while excelling in the accuracy-efficiency trade-off. Specifically, TabularARGN offers significant improvements in computational efficiency, with faster training and inference times, without sacrificing accuracy. This balance is particularly important for real-world applications, where datasets are often characterized by high dimensionality, mixed feature types, and diverse sequential structures.

\newpage

\section*{Acknowledgements}
We sincerely thank Klaudius Kalcher, Roland Boubela, Felix Dorrek, and Thomas Gamauf, whose foundational ideas and early contributions were instrumental in shaping many of the key architectural components of TabularARGN and guiding its initial implementation.
We also wish to thank Radu Rogojanu, Lukasz Kolodziejczyk, Michael Druk, Shuang Wu, Andr\'e Jonasson, Dmitry Aminev, Peter Bognar, Victoria Labmayr, Manuel Pasieka, Daniel Soukup, Anastasios Tsourtis, Jo\~ao Vidigal, Kenan Agyel, Bruno Almeida, Jan Valendin, and M{\"u}rsel Ta\c{s}g{\i}n. Their collective efforts have significantly contributed to the development and refinement of TabularARGN, enhancing its ability to generate high-quality synthetic data while ensuring robust protection against information leakage. Their remarkable work has made TabularARGN an efficient, fast, robust, flexible, and user-friendly solution for tabular synthetic data generation.

\bibliography{paper}
\bibliographystyle{icml2025}

\newpage
\appendix
\onecolumn


\section{Encoding Strategies}
\label{A:encoding}
The TabularARGN framework supports a wide range of data types, including categorical, numerical, datetime, string, and geospatial. As the TabularARGN models exclusively operate on categorical columns, all other data types are converted into one or more categorical sub-columns using specific encoding strategies.

\textbf{Categorical columns} require no modification, as they are inherently categorical. However, missing or empty values are treated as separate categories and are faithfully reproduced in the synthetic data.

\textbf{Numerical Columns}: TabularARGN has a pre-defined logic in place to automatically select one of three discretization mechanisms for numerical columns: Numeric-Discrete, Numeric-Binned, or Numeric-Digit.
\begin{itemize}
    \item Numeric-Discrete: Treats each unique numerical value as a distinct category.
    \item Numeric-Binned: Groups numerical values into a maximum of 100 intervals, treating each interval as a category. During generation, values are randomly and uniformly sampled from the selected interval. An additional category is added to indicate missing values, ensuring their reappearance in the synthetic data.
    \item Numeric-Digit: Splits floating-point values into individual digit-based sub-columns. For instance, a column containing values like 12.3 is encoded into three categorical sub-columns representing digits 1, 2, and 3, with additional indicator sub-columns for signs and missing values (if applicable).
\end{itemize}

\textbf{Datetime Columns}: TabularARGN accepts a wide variety of datetime formats. These columns are split into sub-columns for year, month, day, hour, minute, second, and, if applicable, milliseconds which all are treated as categorical columns.

For sequential tables, TabularARGN also provides a relative datetime encoding type, which translates original data into intra-record intervals. The first entry in a sequence is normalized to time $t=0$, and subsequent entries are represented as the time difference relative to the first entry.

\textbf{Character encoding} is applicable for columns containing strings of small to medium length. Each string is split into individual characters, resulting in as many sub-columns as the maximum string length in the column. Each character sub-column is treated categorically.

\textbf{Geospatial columns}: TabularARGN supports geospatial data specified as latitude and longitude. Internally, the two numeric values are converted into a hierarchical character sequence of quad-tiles\footnote{\url{https://wiki.openstreetmap.org/wiki/QuadTiles}}, which are encoded as multiple categorical sub-columns, each increasing in spatial specificity.

\section{Heuristics and Layer Sizes}
\label{A:heuristics}

The table below provides an overview of the layer sizes and dimensions for the key components of the TabularARGN models, including the embedding layers, regressor layers, context processors, and history compressors.

\begin{table}[h]
\centering
\begin{tabular}{lllccc}
Layer & Flat & Sequential & Heuristic \\
\toprule
Embeddings                & yes & yes & $3 \cdot (d_{\text{in}})^{\mathstrut 0.25} $ \\[0.2cm]
Regressors                & yes & yes & $16 \cdot \text{max}(1, \text{ln}(d_{\text{in}})) $ \\[0.2cm]
Context Compressor           & no & yes & $64 \cdot \text{max}(1, \text{ln}(d_{\text{ctx}})) $ \\[0.2cm]
History Compressor               & no & yes & $32 \cdot \text{max}(1, \text{ln}(d_{\text{tgt}} \cdot s_{\text{q50}} )) $ \\
\bottomrule
\end{tabular}
\caption{Heuristic for calculating the size of the model layers. $d_{\text{in}}$ is the cardinality of the input sub-column. $d_{\text{ctx}}$ is the full dimension of the concatenated flat context column embeddings. $d_{\text{tgt}}$ is the full dimension of the concatenated sequential target column embeddings. $s_{\text{q50}}$ is the median sequence length for the sequential table.}
\label{tab:heuristics}
\end{table}

\section{Model Training}
\label{A:model-training}

To ensure efficient and robust model training, TabularARGN employs an early stopping mechanism based on validation loss. This mechanism prevents overfitting and reduces unnecessary training time by halting the process when further improvements in validation performance are unlikely.

Validation Loss and Early Stopping: During training, 10\% of the input dataset is split off as a validation set. At the end of each epoch, the model’s validation loss is calculated. If the validation loss does not improve for $N$ consecutive epochs, training is stopped. The default setting for $N$ is 5 epochs. The model weights corresponding to the lowest observed validation loss are retained as the final trained model.

Learning Rate Scheduler: In addition to early stopping, a learning rate scheduler is employed to dynamically adjust the learning rate during training. If the validation loss does not improve for $K$ consecutive epochs, the learning rate is halved to promote finer adjustments in the model's parameters. The default setting for $K$ is 3 epochs.

These mechanisms work together to optimize the training process, ensuring that the model converges efficiently while minimizing the risk of overfitting.

\section{Sequential Model}
\label{A:sequential-table-model}
\textbf{Sequence Length}: The sequential table Tabular ARGN model achieves its flexibility in handling sequences of arbitrary lengths and irregular time steps by incorporating two additional columns into each sequence during preprocessing. These columns are designed to act as a form of positional encoding, providing the model with the necessary context for both training and generation.
\begin{enumerate}
    \item  Sequence Length Column: This column is constant for each sequence, denoting the total length of the corresponding sequence. It remains unchanged across all rows of the sequence and serves as a global identifier for the sequence's size.
    \item  Counting Index Column: This column assigns a unique index to each time step within the sequence, starting at 1 for the first time step and incrementing sequentially up to the sequence length. It acts as a positional marker, allowing the model to identify the relative position within the sequence.
\end{enumerate}
Both columns are treated as numerical features, ensuring compatibility with the model's architecture. During training, these additional columns are fed to the model along with the other features, enabling the model to learn relationships between sequence positions and the observed data.

Generation Process: During generation, the model predicts the sequence length in the first time step - optionally conditioned on a flat context table (if present) - and is kept constant for subsequent time steps. Once the sequence length is determined, the counting index is incremented deterministically for subsequent time steps. This ensures that the model tracks its position within the sequence as it generates synthetic data step by step. Together, these columns provide a structured yet flexible mechanism to handle sequences of varying lengths and irregular spacing, making the model suitable for a wide range of real-world applications, including time-series data and ordered sets.

\textbf{Max sequence window}: Some datasets include sequences with very long spans, which can significantly increase training time. However, in many practical cases, the temporal correlations of interest occur over shorter scales than the full sequence length. To address this, we introduce a tunable max sequence window during the training phase of sequential models.

Training with Max Sequence Window: Instead of training on the full sequence length for each sample, a fixed window of time steps, referred to as the max sequence window, is selected. This allows the model to focus on shorter subsequences. The default max sequence window size is set to 100 time steps but can be adjusted manually to suit the characteristics of the dataset being synthesized.

For each batch, and for every sequence within that batch, a random starting point is selected for the window. This starting point is chosen uniformly, with corrections applied to ensure that time steps near the beginning and end of sequences are not underrepresented. Only the subsequence defined by this max sequence window is used for gradient calculation during training, significantly reducing computational demands for long-sequence datasets.

The generation process remains consistent with the standard approach: the sequence length is predicted in the first time step (optionally conditioned on a flat context table), and the counting index is incremented deterministically as the model generates the sequence. This ensures that the synthetic data faithfully captures the structure of the original dataset, even when the training process leverages shorter max sequence windows.

With a default value of 100, the max sequence window is only relevant for the Berka data set in this paper as the average sequence lengths of all other data sets are below 10.

\section{Metrics}
\label{A:metrics}
The metrics used for the evaluation of synthetic-data quality follow the general approach of \cite{pla21-qa} and are available in a well-maintained and documented open-source GitHub repository \cite{mostlyai-qa}. We include metrics for measuring low-dimensional marginal statistics and a distance-based metric to measure the novelty, i.e.\ privacy of the synthetic data.

\subsection{Low-Order Marginal Statistics}
Low-order marginal statistics are evaluated by comparing univariate (column-wise) distributions and the pairwise correlations between columns. To handle mixed-type data, numerical and DateTime columns are discretized by grouping their values into deciles defined by the original training data, resulting in 10 groups per column (equally sized for the original data). For categorical columns, only the 10 most frequent categories are retained, with the remaining categories disregarded. This approach ensures comparability across data types while focusing on the most significant features of the data.

For each feature, we derive a vector of length 10 from the training (original) data and another from the synthetic data. For numerical and DateTime columns, the vectors represent the frequencies of the groups defined by the original deciles. For categorical columns, the vectors reflect the frequency distribution of the top 10 categories after re-normalization. These feature-specific vectors are denoted as $\mathbf{X}_{\text{trn}}^{(m)}$ and $\mathbf{X}_{\text{syn}}^{(m)}$, corresponding to the training and synthetic data, respectively. $m$ is the feature index, running from 1 to $d$.

The \textbf{univariate accuracy} of column $m$ is then defined as
\begin{equation}
    acc_{\text{univariate}}^{(m)} = \frac{1}{2} \left(1 - \|\mathbf{X}_{\text{trn}}^{(m)} - \mathbf{X}_{\text{syn}}^{(m)}\|_1 \right)
\end{equation}
and the overall univariate accuracy, as reported in the results section, is defined by
\begin{equation}
    acc_{\text{univariate}} = \frac{1}{D} \sum_m^D acc_{\text{univariate}^{(m)}} \; ,
\end{equation}
where $D$ is the number of columns.

For bi-variate metrics, we evaluate the relationships between pairs of columns by constructing normalized contingency tables. These tables capture the joint distribution of two features, $m$ and $n$, allowing us to assess pairwise dependencies.

The contingency table between columns $m$ and $n$ is denoted as $\mathbf{C}_{\text{trn}}^{(m,n)}$ for the training data and $\mathbf{C}_{\text{syn}}^{(m,n)}$ for the synthetic data. Each table has a maximum dimension of 10×10, corresponding to the (discretized) values or the top 10 categories of the two features. For columns with fewer than 10 categories (categorical columns with cardinality \textless 10), the dimensions of the table are reduced accordingly.

Each cell in the table represents the normalized frequency with which a specific combination of categories or discretized values from columns $m$ and $n$ appears in the data. This normalization ensures that the contingency table is comparable across features and datasets, regardless of their absolute scale or size.

The \textbf{bivariate accuracy} of the column pair $m,n$ is defined as
\begin{equation}
    acc^{(m,n)}_{\text{bivariate}} = \frac{1}{2}\left(1 - \|\mathbf{C}_{\text{trn}}^{(m,n)} - \mathbf{C}_{\text{syn}}^{(m,n)}\|_{1,\text{entrywise}}\right) = \frac{1}{2}\left(1 - \sum_i \sum_j = \left|\mathbf{C}_{\text{trn}}^{(m,n)} - \mathbf{C}_{\text{syn}}^{(m,n)}\right|_{i,j}\right)
\end{equation}
and the overall bivariate accuracy, as reported in the results section, is given by
\begin{equation}
    acc_\text{bivariate} \frac{2}{D(D-1)} \sum_{1 \leq m < n \leq D} acc_\text{bivariate}^{(m,n)} \; ,
\end{equation}  
the average of the strictly upper triangle of $acc_\text{bivariate}^{(m,n)}$.

Note that due to sampling noise, both $acc_{\text{univariate}}$ and $acc_{\text{bivariate}}$ cannot reach 1 in practice. The software package reports the theoretical maximum alongside both metrics.

There is no difference in calculating the univariate and bivariate accuracies between flat and sequential data. In both cases, vectors $\mathbf{X}^{(m)}$ and contingency tables $\mathbf{C}^{(m,n)}$ are based on all entries in the columns, irrespective of which data subject they belong to.

The coherence metric, specific to sequential data, evaluates the consistency of relationships between successive time steps or sequence elements. It is conceptually similar to the bi-variate accuracy metric but adapted for sequential datasets. The process is as follows:
\begin{itemize}
    \item For each data subject, we randomly sample two successive sequence elements (time steps) from their sequential data.
    \item These pairs of successive time steps are transformed into a wide-format dataset. To illustrate, consider a sequential dataset of $N$ subjects and original columns $A, B, C$, represented as $K>N$ rows. After processing, the resulting dataset has six columns: $A, A', B, B', C, C'$. The unprimed columns correspond to the first sampled sequence element, the primed columns correspond to the successive sequence element. The number of rows in this wide-format dataset is equal to $N$, irrespective of the sequence lengths in the original dataset.
\end{itemize}
Using this wide-format dataset, we construct contingency tables $\mathbf{C}^{(m,m')}$ for each pair of corresponding unprimed and primed columns $(m,m')$. These tables are normalized and used to calculate the \textbf{coherence metric} for column $m$ as:
\begin{equation}
    acc^{(m,m')}_{\text{coherence}} = \frac{1}{2}\left(1 - \|\mathbf{C}_{\text{trn}}^{(m,m')} - \mathbf{C}_{\text{syn}}^{(m,m')}\|_{1,\text{entrywise}}\right)
\end{equation}
and the overall coherence metric, as reported in the results section
\begin{equation}
    acc_\text{coherence} = \frac{1}{D} \sum_{m}^{D} acc^{(m,m')}_\text{coherence} \; .
\end{equation}

We summarize the \textbf{overall accuracy} of a data set as
\begin{equation}
    \frac{1}{2} \left( acc_\text{univariate} + acc_\text{bivariate} \right)
\end{equation} and
\begin{equation}
    \frac{1}{3} \left( acc_\text{univariate} + acc_\text{bivariate} + acc_\text{coherence} \right)
\end{equation}
for flat and sequential data, respectively.

\subsection{Privacy Metric}
Since our privacy metric relies on distances between samples, each record is mapped into an embedding space. To achieve this, a tabular sample is first converted into a string of values (e.g., \texttt{value\_col1;value\_col2;...;value\_colD}), which is then processed by a pre-trained language model. We opt for \texttt{all-MiniLM-L6-v2}\footnote{\url{https://huggingface.co/sentence-transformers/all-MiniLM-L6-v2/}} as it is a lightweight, compute-efficient universal model. It transforms each string of values into a 384-dimensional embedding space. For sequential data, the string is constructed by concatenating the values of all columns across time steps. For instance, values from time step two are appended to the string containing values from time step one, and so on. For long sequences, the resulting input string is truncated to fit within the model’s context window.


To score the leakage of information from training to synthetic data, we adopt the DCR share metric, as proposed by \citet{pla21-qa}. This metric assesses whether synthetic data points are closer to the training data than to a holdout set of the same size, serving as an indicator of potential overfitting. For each synthetic record, the nearest neighbor is identified within both the training and holdout datasets. The DCR share metric is defined as the fraction of synthetic records whose nearest neighbor lies in the training data. A value close to 0.5 suggests that the generative model has not overfitted and instead samples from the approximate distribution of the original data rather than copying from the training data.

With the sample embeddings denoted as $\text{emb}_i$ and $i$ ranging from $1$ to $N$, nearest neighbor distances are calculated using the L2 norm between embedded representations of synthetic, training, and holdout records. For an embedded synthetic record $\text{emb}_i^\text{(syn)}$, the distance to its nearest neighbor in the training and holdout datasets is computed as:
\begin{equation}
    d_{\text{trn}}^{(i)} = \min_{j \in N_{\text{trn}}} \|\text{emb}_i^{(\text{syn})} - \text{emb}_j^{(\text{trn})}\|_2, \quad
    d_{\text{hold}}^{(i)} = \min_{j \in N_{\text{hold}}} \|\text{emb}_i^{(\text{syn})} - \text{emb}_j^{(\text{hold})}\|_2.
\end{equation}
With the indicator function
\begin{equation}
\mathbb{I}_{\text{trn}}^{(i)} =
\begin{cases} 
1 & \text{if } d_{\text{trn}}^{(i)} < d_{\text{hold}}^{(i)}, \\
0 & \text{if } d_{\text{trn}}^{(i)} > d_{\text{hold}}^{(i)}, \\
0.5 & \text{if } d_{\text{trn}}^{(i)} = d_{\text{hold}}^{(i)},
\end{cases}
\end{equation}
which indicates whether the nearest neighbor of $\text{emb}_i^\text{(syn)}$ is in the training set, we define the \textbf{DCR share} as
\begin{equation}
\text{DCR share} = \frac{1}{N_{\text{syn}}} \sum_{i=1}^{N_{\text{syn}}} \mathbb{I}_{\text{trn}}^{(i)}.
\end{equation}


\section{Experimental setup}
\label{A:experimental-setup}

\textbf{Compute}: We perform almost all experiments on \texttt{AWS g5.2xlarge} instances with 8 vCPUs, 32 GiB system RAM, and one NVIDIA A10G (24 GiB of GPU memory). As a non-deep learning method, UnmaskingTrees is specifically designed to run on CPU and we opted for an \texttt{AWS p3.2xlarge} instance with 8 vCPUs and 61 GiB of system RAM. 

\textbf{Differential Privacy}: Differentially private SGD is implemented using the Opacus library \cite{you21-opacus}. For all runs with DP-SGD, we use a noise multiplier of 1.5, a DP max.~grad.~norm of 1, and a $\delta = 1 \times 10^{-5}$.

\textbf{Data sets}: The \emph{Adult} data set \citep{dua19-adult} has around 48k rows with eight categorical and six numerical features. \emph{ACS-Income} \cite{din21-acsI, flo20-acsII} has about 1.5 million rows, containing 28 categorical and four numerical features—making it roughly 30 times larger and twice as wide as \emph{Adult}, and thus a challenging real-world benchmark. 
\emph{Default} \citep{yeh2009default} contains 30,000 financial records of credit card users labeled by default status, and \emph{Shoppers} \citep{sakar2018online} consists of 12,330 user sessions, capturing browsing behavior and transaction outcomes on an e-commerce platform.
We initially considered an even larger data set with more columns, but all baseline methods were either too slow or required excessive GPU memory. 

For sequential modeling, we present results on a two-table subset (\textit{players} and \textit{fielding}) of the \emph{Baseball} dataset \citep{lahman2023baseball}, the \emph{California} dataset \citep{pace1997sparse} consisting of a table of \textit{households} and a table of \textit{individuals}, and \emph{Berka} dataset \citep{berka1999}, where we only consider the \textit{accounts} and \textit{transactions} tables.
 For \emph{Baseball}, the \textit{players} table (flat context) has about 20 thousand rows, and the sequential \textit{fielding} table has 140 thousand, resulting in an average sequence length of seven. 
With over 600 thousand rows, the \textit{household} table has about $30$ times more records than the \textit{players} table, resulting in almost 1.7 million rows and an average sequence length of 2.8 in the \textit{individuals} table. 
 The \textit{accounts} table has only 4,500 records, but the \textit{transactions} table is notably larger, encompassing around 1 million rows.
 All sequence tables have a comparable number of features: 9 (\textit{Berka}), 11 (\textit{Baseball}) and 15 (\textit{California})

Details on the flat data sets are shown in table \ref{tab:datasets-flat}; details on the sequential data sets, in \ref{tab:datasets-sequence}.

\begin{table}[ht]
\centering
\begin{tabular}{lrrrrr}
\toprule
\textbf{Name} & \textbf{Training Rows} & \textbf{Holdout Rows} & \textbf{Categorical} & \textbf{Numerical} \\
\midrule
Adult       & 24,421   & 24,421   & 8 & 6 \\
ACS-income  & 738,108  & 738,108  & 28 & 4 \\
Default & 15,000 & 15,000 & 10 & 13 \\
Shoppers & 6,165 & 6,165 & 8 & 10 \\  
\bottomrule
\end{tabular}
\caption{Details of flat-data benchmark datasets.}
\label{tab:datasets-flat}
\end{table}

\begin{table}[ht]
\centering
\begin{tabular}{lrrrrrr}
\toprule
\textbf{Dataset} & \textbf{Train Rows} & \textbf{Avg. Train Len.} & \textbf{Holdout Rows} & \textbf{Avg. Holdout Len.} & \textbf{Num. Feat.} & \textbf{Cat. Feat.} \\
\midrule
\multicolumn{7}{l}{\textbf{Sequential Data}} \\
Baseball    & 71,095   & 7.11   & 70,747   & 7.16   & 8  & 3 \\
California  & 844,641  & 2.74   & 846,001  & 2.75   & 15 & 0 \\
Berka & 526,442 & 234 & 529,878 & 236 & 9 & 2 \\
\midrule
\multicolumn{7}{l}{\textbf{Flat Context Data}} \\
Baseball    & 10,000   & -      & 9,878   & -      & 2  & 5 \\
California  & 308,057  & -      & 308,058 & -      & 10 & 0 \\
Berka & 2,250 & - & 2,250 & - & 4 & 1 \\
\bottomrule
\end{tabular}
\caption{Details of sequential-data benchmark datasets including the flat context.}
\label{tab:datasets-sequence}
\end{table}

\textbf{Baselines}: 
The baseline methods included in this study can be grouped depending on whether they process flat tables or sequential tables.

For flat tables, we include CT-GAN, a generative model introduced by \citet{xu19} that applies mode-specific normalization to handle multi-modal data while generating numerical features conditioned on categorical ones. STaSy \citep{kim23-StaSy} builds on score-based generative modeling, incorporating self-paced learning and fine-tuning strategies to improve the stability of denoising score matching. TabSyn \citep{zha24-TabSyn} adopts a VAE to project mixed data types into a common latent space, allowing a diffusion model to better capture the underlying data distribution. Unmasking Trees, proposed by \citet{mcc24-UnmaskingTrees}, is an alternative route to deep learning models. This framework leverages gradient-boosted decision trees to incrementally unmask individual features. 

For sequential tables, we consider REalTabFormer \citep{sol23-Realtabformer}, a transformer-based framework that generates flat records autoregressively using a GPT-2 model while employing a sequence-to-sequence approach to synthesize child tables conditioned on parent data. SDV \citep{pat16-sdv} takes a different approach by iteratively modeling relationships across a relational database, enabling flexible data synthesis. ClavaDDPM \citep{pan24-clavaddpm} introduces a cluster-guided diffusion mechanism that captures inter-table dependencies via latent representations, improving the modeling of complex multi-relational structures. Finally, RC-TGAN \citep{gue23-rctgan} conditions the generation of child table rows on parent data, extending its modeling capacity to high-order relationships across multiple interconnected tables.

\textbf{Code repositories}: Table \ref{tab:methods_repos} lists the repositories containing implementations of all methods used for the benchmarks in this paper. Throughout this benchmark, we use the default settings of \emph{TabularARGN} as implemented in x.  \emph{TabSyn} and \emph{STaSy} are implemented in \url{https://github.com/amazon-science/tabsyn}. For implementation details of these baselines, we refer to appendix G.2 of \cite{zha24-TabSyn}.

\begin{table}[ht]
\centering
\begin{tabular}{ll}
\toprule
\textbf{Method} & \textbf{Repository} \\
\midrule
\emph{TabularARGN} & \url{https://github.com/mostly-ai/mostlyai-engine} \\
\emph{TabSyn} & \url{https://github.com/amazon-science/tabsyn} \\
\emph{STaSY} & \url{https://github.com/amazon-science/tabsyn} \\
\emph{TabMT} & \url{https://github.com/fangliancheng/TabDAR}$^*$ \\
\emph{CT-GAN} & \url{https://github.com/vanderschaarlab/synthcity} \\
\emph{UnmaskingTrees} & \url{https://github.com/calvinmccarter/unmasking-trees} \\
\emph{SDV} & \url{https://github.com/sdv-dev/SDV} \\
\emph{RC-TGAN} & \url{https://github.com/croesuslab/RCTGAN} \\
\emph{REaLTabFormer} & \url{https://github.com/worldbank/REaLTabFormer} \\
\emph{ClavaDDPM} & \url{https://github.com/weipang142857/ClavaDDPM} \\
\bottomrule
\end{tabular}
\caption{List of benchmark methods and their repositories. $^*$As of January 2025, the TabDAR repo is not publicly available. We thank \citep{zha24-TabDar} for providing their code upfront.}
\label{tab:methods_repos}
\end{table}

\section{Further Results}
\label{A:further-results}

This section provides a comprehensive overview of results not included in the main text. Specifically, we report results on additional data sets (see description in \ref{A:experimental-setup}) for both flat and sequential tables. 

\begin{figure*}
    \centering
    \includegraphics[width=1.0\textwidth]{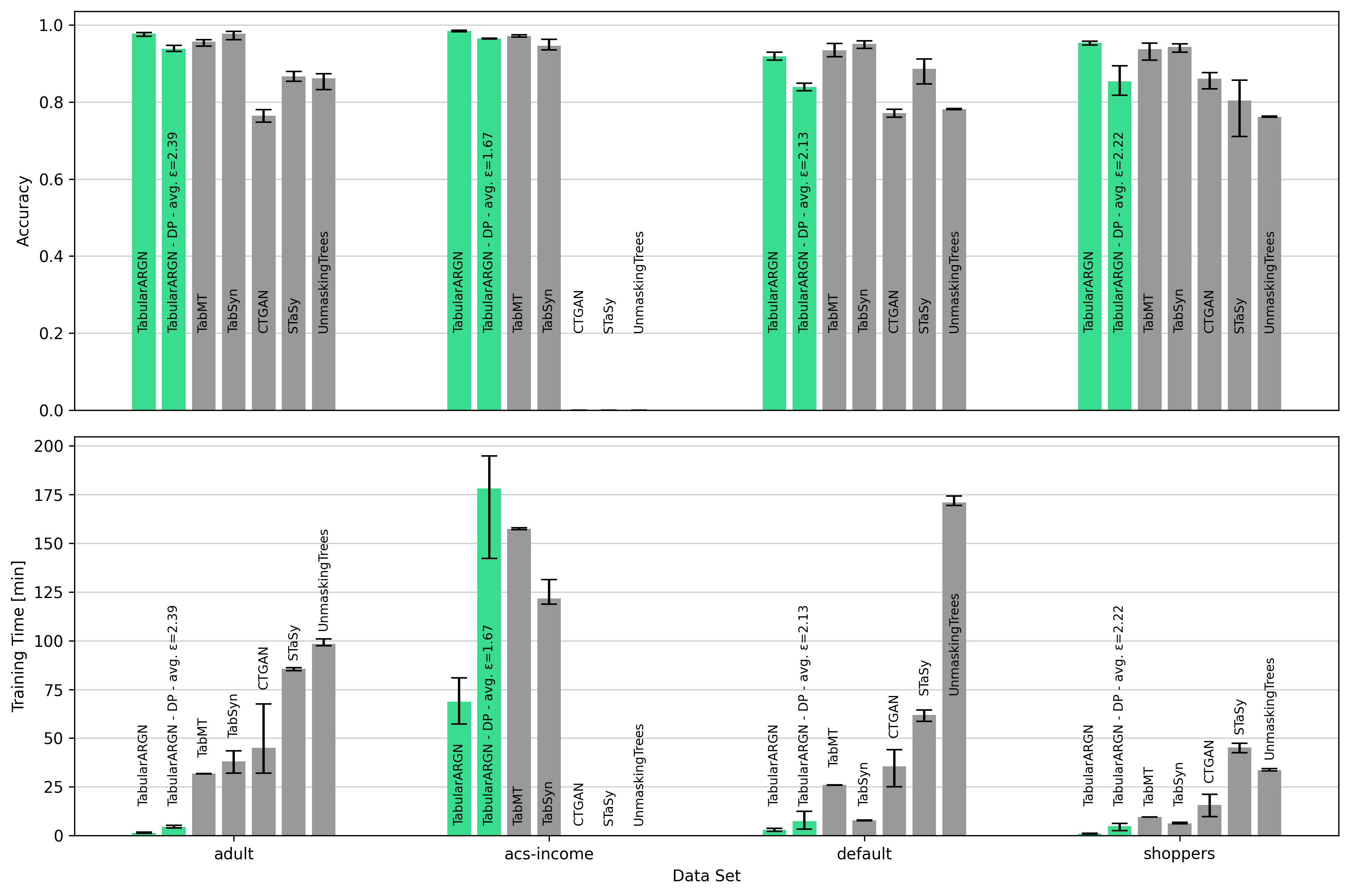}
    \caption{Accuracy (top row) and Training Time (bottom row) for the flat datasets \emph{Adult}, \emph{ACS-Income}, \emph{Default}, and \emph{Shoppers} and various generative models - TabularARGN, TabularARGN with DP, TabMT, TabSyn, CTGAN, STaSy, and UnmaskingTree. Reported values are averages over five full training and generation runs, error bars indicate the minimum and maximum values. As UnmaskingTrees is a non-Deep Learning approach, it is run on CPU. For comparison, Training of TabularARGN on the \emph{Adult} with a CPU takes approximately 2.5 minutes.}
    \label{fig:results_flat}
\end{figure*}

Similar to the \emph{Adult} data set, TabularARGN performs on par concerning accuracy on the \emph{ACS-Income}, \emph{Default}, and \emph{Shoppers} data set with SOTA models (see Fig.~\ref{fig:results_flat}). For some data sets, it marginally surpasses them. Throughout all data sets, the training time of TabularARGN is the smallest, often by a large margin. The strongest competitor in the accuracy and accuracy-efficiency trade-off is TabSyn. Applying DP-SGD increases training time and reduces accuracy. The drop in accuracy is smaller for data sets with a large sample size as the signal of individual records is better protected. 

The \emph{ACS-Income} data set has considerably more rows (approx.\ 700k in the training split) than all the other ones. This not only increases training time but makes running a subset of models difficult. UnmaskingTrees caused an OOM error on a machine with 61 GiB of RAM and CTGAN did not finish after 16 hours of training. STaSy did finish but resulted in very low accuracy numbers. We refrain from publishing these numbers as they might stem from issues other than the method itself.

For the \emph{Adult} dataset, we also ran the metrics used in the TabSyn benchmark (see Tab.\ \ref{tab:tabsyn-results}), including column-wise density and pair-wise column correlation\footnote{\url{https://docs.sdv.dev/sdmetrics}}, as well as the alpha precision and beta recall \cite{alaa2022faithfulsyntheticdatasamplelevel}. As the first two metrics are conceptually similar to those used in our benchmark, they yield results that align closely with the picture described in the results section \ref{sec:results-flat}. Alpha precision and beta recall measure the higher-order fidelity of the synthetic data.

\begin{table}[H]
\centering
\begin{tabular}{lrrrr}
\toprule
\textbf{Model} & \textbf{Marginal} & \textbf{Joint} & \textbf{$\alpha$-Precision} $\uparrow$ & \textbf{$\beta$-Recall} $\downarrow$ \\
\midrule
TabularARGN & 98.7 & 96.5 & 98.7 & 45.6 \\
TabSyn & 99.3 & 96.7 & 99.4 & 46.5 \\
TabMT & 97.5 & 94.6 & 98.6 & 49.2 \\
STaSy & 90.8 & 87.9 & 88.4 & 36.7 \\
CTGAN & 79.68 & 77.07 & 50.63 & 10.2 \\
\bottomrule
\end{tabular}
\caption{SDV \cite{pat16-sdv} and high-order metrics \cite{alaa2022faithfulsyntheticdatasamplelevel}.}
\label{tab:tabsyn-results}
\end{table}

\begin{figure*}
    \centering
    \includegraphics[width=1.0\textwidth]{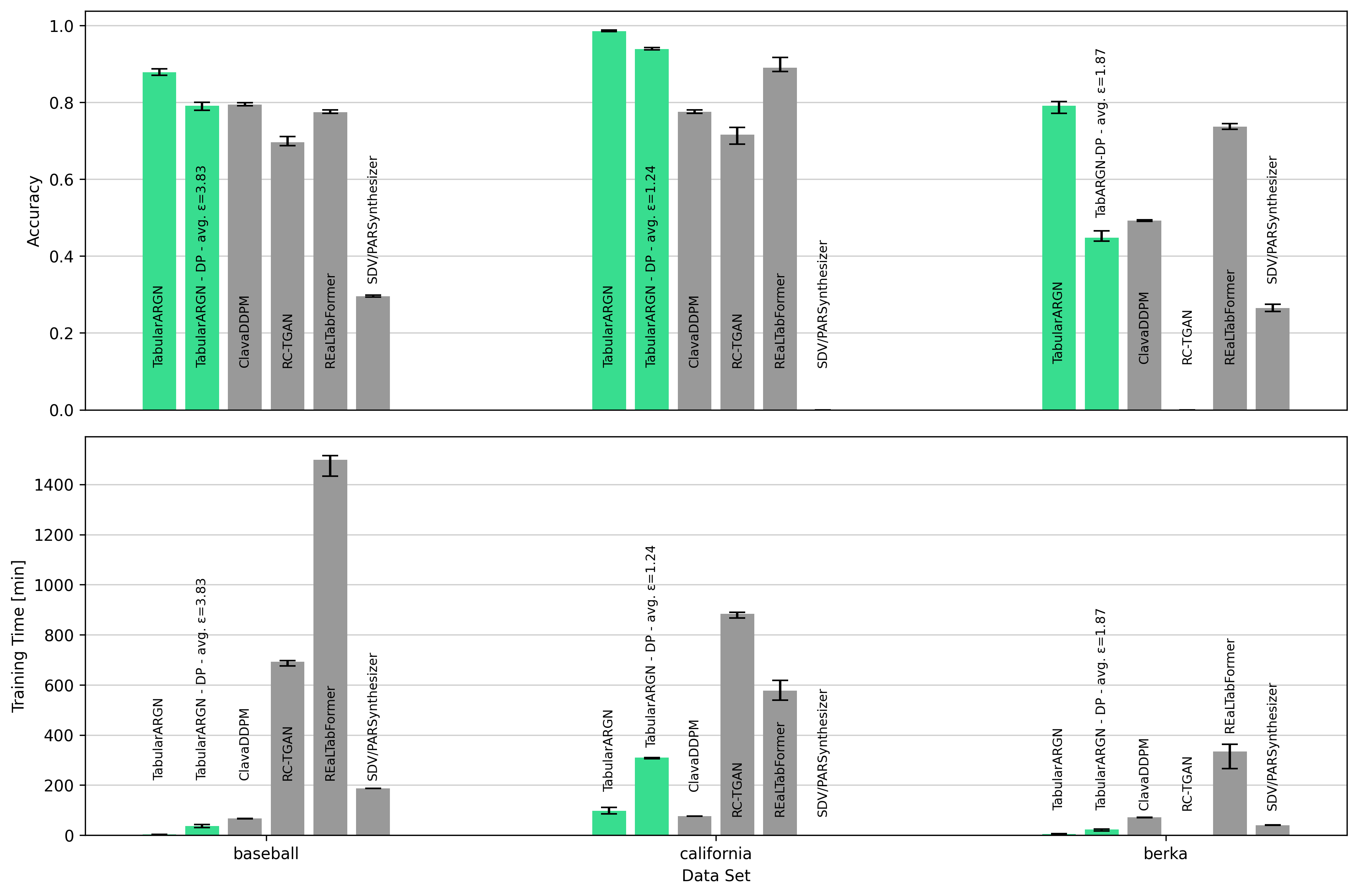}
    \caption{Accuracy (top row) and Training Time (bottom row) for the sequential datasets \emph{Baseball}, \emph{California}, and \emph{Berka} and various generative models - TabularARGN, TabularARGN with DP, ClavaDDPM, RCTGAN, RealTabFormer, and SDV/ParSynthesizer. Reported values are averages over at least three full training and generation runs, error bars indicate the minimum and maximum values. RCTGAN and SDV/ParSynthesizer fail to train on the Berka and California datasets, respectively, due to out-of-memory (OOM) issues.}
    \label{fig:results_seq}
\end{figure*}

For sequential data (see Fig.~\ref{fig:results_seq}) TabularARGN achieves the highest accuracy values across all methods. In terms of runtime, it outperforms almost all of them with the only exception of ClavaDDPM for the \emph{California} data set. However, ClavaDDPM lags in accuracy underpinning the strong accuracy-efficiency trade-off of TabularARGN. RC-TGAN fails on the \emph{Berka} and SDV on the \emph{California} data set with an OOM error.

Applying DP-SGD has the same effect on sequential as it has on flat data. The by far strongest decrease in accuracy happens for the \emph{Berka} data set which has only 2250 training subjects and the longest sequences of all benchmark data sets.

We summarize the key results from Section 4 in Table~\ref{tab:flat_results} for flat datasets and Table~\ref{tab:sequential_results}, presenting a detailed evaluation of synthetic data quality across all relevant metrics. 
For TabularARGN with DP (TabularARGN-DP) we report the average privacy budgets $\epsilon$ in Table~\ref{tab:avg_epsilon}.
The first table reports accuracy, univariate accuracy, bivariate accuracy, coherence, DCR share, and generation time. 
In the second table we add coherence, a metric that assesses the quality of sequential tables specifically.
These metrics offer a comprehensive assessment of the models' ability to capture statistical dependencies and structural patterns in both flat and sequential tabular data.

\begin{table*}[t]
\centering
\resizebox{\textwidth}{!}{
\begin{tabular}{llcccc}
\toprule
Metric & Model & Adult & Default & Shoppers & ACS-Income \\
\midrule
\multirow{7}{*}{Accuracy}  
& TabMT & 0.96 [0.95, 0.96] & 0.93 [0.92, 0.95] & 0.94 [0.91, 0.95] & 0.97 [0.97, 0.97] \\
& TabSyn & \textbf{0.98} [0.96, 0.98] & \textbf{0.95} [0.94, 0.96] & 0.94 [0.93, 0.95] & 0.95 [0.94, 0.96] \\
& STaSy & 0.87 [0.85, 0.88] & 0.89 [0.85, 0.91] & 0.80 [0.71, 0.86] & - \\
& CTGAN & 0.77 [0.75, 0.78] & 0.77 [0.76, 0.78] & 0.86 [0.83, 0.88] & - \\
& Unmasking Trees & 0.86 [0.83, 0.87] & 0.78 [0.78, 0.78] & 0.76 [0.76, 0.76] & - \\
& TabularARGN & \textbf{0.98} [0.97, 0.98] & 0.92 [0.91, 0.93] & \textbf{0.95} [0.95, 0.95] & \textbf{0.98} [0.98, 0.98] \\
& TabularARGN-DP & 0.94 [0.93, 0.95] & 0.84 [0.83, 0.85] & 0.85 [0.82, 0.89] & 0.96 [0.96, 0.96] \\
\midrule
\multirow{7}{*}{Univariate Accuracy}  
& TabMT & 0.97 [0.96, 0.97] & 0.95 [0.94, 0.97] & 0.95 [0.93, 0.97] & \textbf{0.98} [0.98, 0.98] \\
& TabSyn & \textbf{0.99} [0.98, 0.99] & \textbf{0.97} [0.95, 0.98] & 0.96 [0.95, 0.97] & 0.96 [0.95, 0.97] \\
& STaSy & 0.90 [0.88, 0.90] & 0.92 [0.88, 0.94] & 0.85 [0.77, 0.89] & - \\
& CTGAN & 0.83 [0.82, 0.84] & 0.83 [0.82, 0.84] & 0.90 [0.88, 0.91] & - \\
& Unmasking Trees & 0.90 [0.87, 0.91] & 0.83 [0.83, 0.83] & 0.82 [0.82, 0.82] & - \\
& TabularARGN & \textbf{0.99} [0.98, 0.99] & 0.94 [0.93, 0.95] & \textbf{0.97} [0.96, 0.97] & \textbf{0.99} [0.99, 0.99] \\
& TabularARGN-DP & 0.96 [0.95, 0.97] & 0.91 [0.90, 0.91] & 0.90 [0.87, 0.93] & 0.98 [0.97, 0.98] \\
\midrule
\multirow{7}{*}{Bivariate Accuracy}  
& TabMT & 0.94 [0.93, 0.95] & 0.92 [0.90, 0.94] & 0.92 [0.89, 0.94] & 0.96 [0.96, 0.97] \\
& TabSyn & \textbf{0.97} [0.95, 0.98] & \textbf{0.94} [0.92, 0.94] & 0.93 [0.91, 0.94] & 0.93 [0.92, 0.95] \\
& STaSy & 0.84 [0.82, 0.85] & 0.86 [0.81, 0.89] & 0.76 [0.65, 0.83] & - \\
& CTGAN & 0.70 [0.68, 0.72] & 0.71 [0.70, 0.72] & 0.82 [0.79, 0.84] & - \\
& Unmasking Trees & 0.83 [0.79, 0.84] & 0.73 [0.73, 0.74] & 0.71 [0.70, 0.71] & - \\
& TabularARGN & \textbf{0.97} [0.96, 0.97] & 0.90 [0.89, 0.91] & \textbf{0.94} [0.93, 0.94] & \textbf{0.98} [0.98, 0.98] \\
& TabularARGN-DP & 0.92 [0.91, 0.93] & 0.77 [0.76, 0.78] & 0.81 [0.77, 0.86] & 0.95 [0.95, 0.96] \\
\midrule
\multirow{7}{*}{DCR Share}  
& TabMT & 0.51 [0.51, 0.51] & 0.50 [0.50, 0.50] & 0.50 [0.50, 0.50] & 0.50 [0.50, 0.50] \\
& TabSyn & 0.50 [0.50, 0.50] & 0.50 [0.50, 0.50] & 0.50 [0.50, 0.50] & 0.50 [0.50, 0.50] \\
& STaSy & 0.50 [0.50, 0.50] & 0.50 [0.50, 0.50] & 0.50 [0.50, 0.50] & - \\
& CTGAN & 0.50 [0.50, 0.50] & 0.50 [0.50, 0.50] & 0.50 [0.50, 0.50] & - \\
& Unmasking Trees & 0.50 [0.50, 0.50] & 0.50 [0.50, 0.50] & 0.50 [0.50, 0.50] & - \\
& TabularARGN & 0.51 [0.51, 0.51] & 0.50 [0.50, 0.50] & 0.50 [0.50, 0.50] & 0.50 [0.50, 0.50] \\
& TabularARGN-DP & 0.50 [0.50, 0.50] & 0.50 [0.50, 0.50] & 0.50 [0.50, 0.50] & 0.50 [0.50, 0.50] \\
\midrule
\multirow{7}{*}{Generation Time (s)}  
& TabMT & 10.91 [10.90, 10.92] & 12.41 [12.38, 12.49] & 7.28 [7.26, 7.30] & - \\
& TabSyn & 7.21 [7.00, 7.27] & 6.33 [6.27, 6.57] & 4.52 [4.52, 4.53] & - \\
& STaSy & 16.83 [14.42, 21.87] & 10.10 [9.86, 10.62] & 8.05 [7.44, 9.74] & - \\
& CTGAN & \textbf{0.15} [0.15, 0.16] & \textbf{0.17} [0.16, 0.17] & \textbf{0.09} [0.09, 0.10] & - \\
& Unmasking Trees & 641.10 [634.63, 646.02] & 663.34 [652.94, 677.04] & 187.09 [181.14, 197.17] & - \\
& TabularARGN & 0.36 [0.33, 0.42] & 0.77 [0.76, 0.78] & 0.30 [0.30, 0.31] & 15.00 [15.00, 15.00] \\
& TabularARGN-DP & 0.37 [0.34, 0.43] & 0.78 [0.77, 0.78] & 0.31 [0.31, 0.32] & \textbf{14.80} [14.67, 14.94] \\
\bottomrule
\end{tabular}
}
\caption{Evaluation of synthetic data quality across flat datasets, averaged over five independent runs. For each metric, we present the mean, minimum, and maximum values as Mean[Min, Max]. The highest mean accuracy for each dataset is highlighted in bold.
Similarly, the most efficient generation times are highlighted. Missing generation times for the ACS-Income data set correspond to generation runs that yielded an Out-of-Memory error in GPU. We do report values for Unmasking Trees as it is designed for training and inference on CPU. Regarding DCR Share, values close to 0.5 indicate the model is not overfitted. Missing values in fields other than Generation Time correspond to experiments that yielded an Out-of-Memory error or where a single experiment ran for more than 16 hrs.}
\label{tab:flat_results}
\end{table*}

\begin{table*}[t]
\centering
\resizebox{\textwidth}{!}{
\begin{tabular}{llccc}
\toprule
Metric & Model & Baseball & Berka & California \\
\midrule

\multirow{6}{*}{Accuracy}  
& ClavaDDPM & 0.79 [0.79, 0.80] & 0.49 [0.49, 0.49] & 0.77 [0.77, 0.78] \\
& REaLTabFormer & 0.77 [0.77, 0.78] & 0.74 [0.73, 0.75] & 0.89 [0.88, 0.92] \\
& RC-TGAN & 0.70 [0.69, 0.71] & - & 0.72 [0.69, 0.74] \\
& SDV/PARSynthesizer & 0.30 [0.29, 0.30] & 0.27 [0.26, 0.28] & - \\
& TabularARGN & \textbf{0.88} [0.87, 0.89] & \textbf{0.79} [0.77, 0.80] & \textbf{0.99} [0.98, 0.99] \\
& TabularARGN-DP & 0.79 [0.78, 0.80] & 0.45 [0.44, 0.47] & 0.94 [0.94, 0.94] \\
\midrule

\multirow{6}{*}{Univariate Accuracy}  
& ClavaDDPM & 0.88 [0.88, 0.88] & 0.63 [0.63, 0.63] & 0.83 [0.83, 0.83] \\
& REaLTabFormer & 0.83 [0.82, 0.84] & 0.81 [0.80, 0.81] & 0.92 [0.91, 0.93] \\
& RC-TGAN & 0.80 [0.79, 0.81] & - & 0.80 [0.79, 0.81] \\
& SDV/PARSynthesizer & 0.40 [0.40, 0.41] & 0.34 [0.33, 0.35] & - \\
& TabularARGN & \textbf{0.95} [0.94, 0.96] & \textbf{0.87} [0.85, 0.88] & \textbf{0.99} [0.98, 0.99] \\
& TabularARGN-DP & 0.92 [0.92, 0.93] & 0.56 [0.55, 0.59] & 0.97 [0.97, 0.97] \\
\midrule

\multirow{6}{*}{Bivariate Accuracy}  
& ClavaDDPM & 0.84 [0.84, 0.85] & 0.63 [0.63, 0.64] & 0.80 [0.80, 0.80] \\
& REaLTabFormer & 0.70 [0.69, 0.71] & 0.63 [0.62, 0.64] & 0.90 [0.89, 0.91] \\
& RC-TGAN & 0.71 [0.70, 0.72] & - & 0.71 [0.70, 0.72] \\
& SDV/PARSynthesizer & 0.28 [0.27, 0.29] & 0.23 [0.22, 0.24] & - \\
& TabularARGN & \textbf{0.77} [0.76, 0.78] & \textbf{0.68} [0.66, 0.69] & \textbf{0.98} [0.98, 0.98] \\
& TabularARGN-DP & 0.74 [0.73, 0.74] & 0.36 [0.35, 0.38] & 0.95 [0.95, 0.95] \\
\midrule

\multirow{6}{*}{DCR Share}  
& ClavaDDPM & 0.50 [0.50, 0.50] & 0.50 [0.49, 0.50] & 0.50 [0.50, 0.50] \\
& REaLTabFormer & 0.50 [0.50, 0.50] & 0.52 [0.51, 0.53] & 0.50 [0.50, 0.50] \\
& RC-TGAN & 0.49 [0.47, 0.51] & - & 0.50 [0.50, 0.50] \\
& SDV/PARSynthesizer & 0.50 [0.50, 0.50] & 0.51 [0.51, 0.52] & - \\
& TabularARGN & 0.50 [0.49, 0.50] & 0.50 [0.49, 0.50] & 0.50 [0.50, 0.50] \\
& TabularARGN-DP & 0.50 [0.50, 0.50] & 0.50 [0.50, 0.51] & 0.50 [0.50, 0.50] \\
\midrule

\multirow{6}{*}{Coherence}  
& ClavaDDPM & 0.66 [0.66, 0.67] & 0.38 [0.38, 0.39] & 0.70 [0.70, 0.70] \\
& REaLTabFormer & 0.79 [0.79, 0.79] & 0.76 [0.75, 0.76] & 0.85 [0.84, 0.85] \\
& RC-TGAN & 0.59 [0.58, 0.61] & - & 0.59 [0.58, 0.61] \\
& SDV/PARSynthesizer & 0.20 [0.20, 0.20] & 0.23 [0.22, 0.24] & - \\
& TabularARGN & \textbf{0.91} [0.90, 0.92] & \textbf{0.82} [0.81, 0.83] & \textbf{0.98} [0.98, 0.98] \\
& TabularARGN-DP & 0.72 [0.72, 0.73] & 0.42 [0.41, 0.43] & 0.89 [0.88, 0.89] \\
\midrule

\multirow{6}{*}{Generation Time}  
& ClavaDDPM & 196.05 [192.76, 197.87] & 1299.56 [1269.92, 1319.62] & 2419.92 [2408.43, 2434.85] \\
& REaLTabFormer & 1058.41 [1044.54, 1091.67] & 17169.53 [15366.52, 18395.99] & 23950.47 [16713.84, 27232.35] \\
& RC-TGAN & \textbf{16.69} [12.30, 21.92] & - & \textbf{249.29} [239.07, 259.08] \\
& SDV/PARSynthesizer & 1010.61 [984.48, 1044.59] & 1162.28 [881.63, 1774.27] & - \\
& TabularARGN & 48.55 [48.29, 48.99] & \textbf{64.91} [64.10, 66.81] & 520.21 [519.57, 521.01] \\
& TabularARGN-DP & 35.17 [35.08, 35.33] & 70.30 [68.46, 71.69] & 620.31 [618.81, 622.32] \\

\bottomrule
\end{tabular}
}
\caption{Evaluation of synthetic data quality across sequential datasets, averaged over five independent runs. For each metric, we present the mean, minimum, and maximum values as Mean[Min, Max]. The highest mean accuracy for each dataset is highlighted in bold. Similarly, the most efficient generation times are highlighted. Regarding DCR Share, values close to 0.5 indicate the model is not overfitted. Missing values correspond to experiments that yielded an Out-of-Memory error or where a single experiment ran for more than 16 hrs.}
\label{tab:sequential_results}
\end{table*}

\begin{table}[t]
\centering
\begin{tabular}{llcc}
\toprule
Table Type & Dataset & Average $\epsilon$ & Min/Max $\epsilon$\\
\midrule
Flat & ACS-Income & 1.670 & 1.44/1.78 \\
Flat & Adult & 2.386 & 2.11/2.59 \\
Flat & Default & 2.130 & 1.32/2.92 \\
Flat & Shoppers & 2.224 & 1.52/2.6 \\
\midrule
Sequential & Baseball & 3.826 & 3.43/4.18 \\
Sequential & Berka & 1.866 & 1.59/1.94 \\
Sequential & California & 1.238 & 1.23/1.24 \\
\bottomrule
\end{tabular}
\caption{Average (over five independent runs) and Min/Max $\epsilon$ values for differentially private TabularARGN-DP models across datasets. We report the $\epsilon$ values reached at model convergence.}
\label{tab:avg_epsilon}
\end{table}


\end{document}